\documentclass[11pt,a4paper,table]{article}
\usepackage[hyperref]{acl2021}
\usepackage{times}
\usepackage{latexsym}
\usepackage{graphicx}
\usepackage{amssymb}

\usepackage{microtype}

\aclfinalcopy 

\graphicspath{ {./images/} }

\title{How Should Agents Ask Questions For Situated Learning? \\ An Annotated Dialogue Corpus}

\author{Felix Gervits$^1$, Antonio Roque$^2$, Gordon Briggs$^3$, Matthias Scheutz$^2$, Matthew Marge$^1$\\
$^1$U.S. Army Research Laboratory, Adelphi, MD 20783\\
$^2$Tufts University, Medford, MA 02155\\
$^3$U.S. Naval Research Laboratory, Washington, DC 20375\\
{\normalsize \texttt{\{felix.gervits, matthew.r.marge\}.civ@mail.mil,}}\\ 
{\normalsize \texttt{\{antonio.roque, matthias.scheutz\}@tufts.edu}}\\
{\normalsize \texttt{gordon.briggs@nrl.navy.mil}}
}

\date{}

\begin{document}
\maketitle
\begin{abstract}
Intelligent agents that are confronted with novel concepts in situated environments will need to ask their human teammates questions to learn about the physical world. To better understand this problem, we need data about asking questions in situated task-based interactions. To this end, we present the Human-Robot Dialogue Learning (HuRDL) Corpus - a novel dialogue corpus collected in an online interactive virtual environment in which human participants play the role of a robot performing a collaborative tool-organization task. We describe the corpus data and a corresponding annotation scheme to offer insight into the form and content of questions that humans ask to facilitate learning in a situated environment. We provide the corpus as an empirically-grounded resource for improving question generation in situated intelligent agents. 
\end{abstract}

\section{Introduction}
\label{sec:introduction}

Situated interaction is an area of interest to the Dialogue Systems community \cite{bohusSigdial2019}, with recent papers investigating aspects of language interaction in situated environments both empirically and computationally 
\cite{gervits-EtAl:2020:sigdial,gupta-EtAl:2019:W19-59,kalpakchi-boye:2019:W19-59,kleingarn-EtAl:2019:W19-59}. This topic is critical for the development of technologies that interact with humans in real and virtual environments, including automated vehicles, smart home appliances, robots, and others. Situated agents are typically deployed in open-world environments and possess multiple sensory modalities, so a critical challenge involves enabling such agents to manage uncertainty across modalities and to learn about unfamiliar concepts. By engaging in dialogue with human interlocutors, these challenges can be addressed through effective clarification requests \cite{chernova2014robot}. However, it is not clear what form these clarifications need to take to be most effective. Building on previous corpus-based methods \cite{attari-heckmann-schlangen:2019:W19-59,ginzburg-EtAl:2019:W19-59,gupta-EtAl:2020:sigdial,fuscone-favre-prvot:2020:sigdial,pmlr-v100-thomason20a}, we address this question through the development of a human-human corpus of a collaborative pick-and-place task, and a corresponding annotation scheme of question form and content. The underlying assumption is that the kinds of questions people ask in this task provide good empirical support for indicators to guide agent questions in similar domains, and will inspire approaches for automated question generation moving forward.  

\section{Background}
\label{sec:background}
Prior work has investigated misunderstanding and clarification in human dialogue \cite{schegloff1977preference,clark1996using,marge2015miscommunication,paek2003toward}. In one such analysis, \citet{purver2003means} proposed a scheme of clarification request forms that was applied to a 150,000 word subset of the British National Corpus and shown to cover 99\% of the sub-corpus. While this scheme was applied to a dialogue system for automated clarification generation \cite{purver2004clarie,purver2006clarie}, it has been criticized for being too general and not accounting for certain types of phenomena such as pragmatic uncertainty \cite{rieser2005implications}. Another classification scheme for clarification requests was introduced by \citet{rodriguez2004form}, building on \citet{schlangen2004causes}'s categorization of clarification causes. This scheme showed good coverage when applied to a data set of 22 dialogues from the Bielefeld corpus of task-oriented dialogue. While these analyses are useful for furthering our understanding of clarification requests, prior schemes did not consider situated domains with high degrees of uncertainty. Moreover, they mainly focused on the \textit{form and function} rather than the \textit{content} of clarifications. The content information reflects the particular type of uncertainty that the agent experienced, and tracking that uncertainty helps to inform how agents can use questions (and answers) to manage that uncertainty. 

The contributions of the current work are the following: (1) the presentation of a new annotated corpus, which has been made available for research purposes\footnote{The corpus and additional details can be found at \url{https://github.com/USArmyResearchLab/ARL-HuRDL}.}; (2) an annotation scheme that extends prior schemes to domains involving situated interaction, and also accounts for clarification requests generated to reduce uncertainty across modalities, such as visual feature clarification; (3) an analysis of the corpus including the distribution of categories from our scheme, along with a discussion of how these results can be used to improve the learning capabilities of situated dialogue agents. 

\section{Corpus Collection}
\label{sec:task}

The HuRDL corpus task was designed to investigate how agents can effectively generate questions to clarify uncertainty across multiple modalities in a task-based interaction. The task domain was designed to naturally present participants with novel concepts and procedural knowledge that they needed to learn; in doing so, they would need to use a variety of question types. 

\begin{figure}[t]
	\centering
	\includegraphics[width=\linewidth]{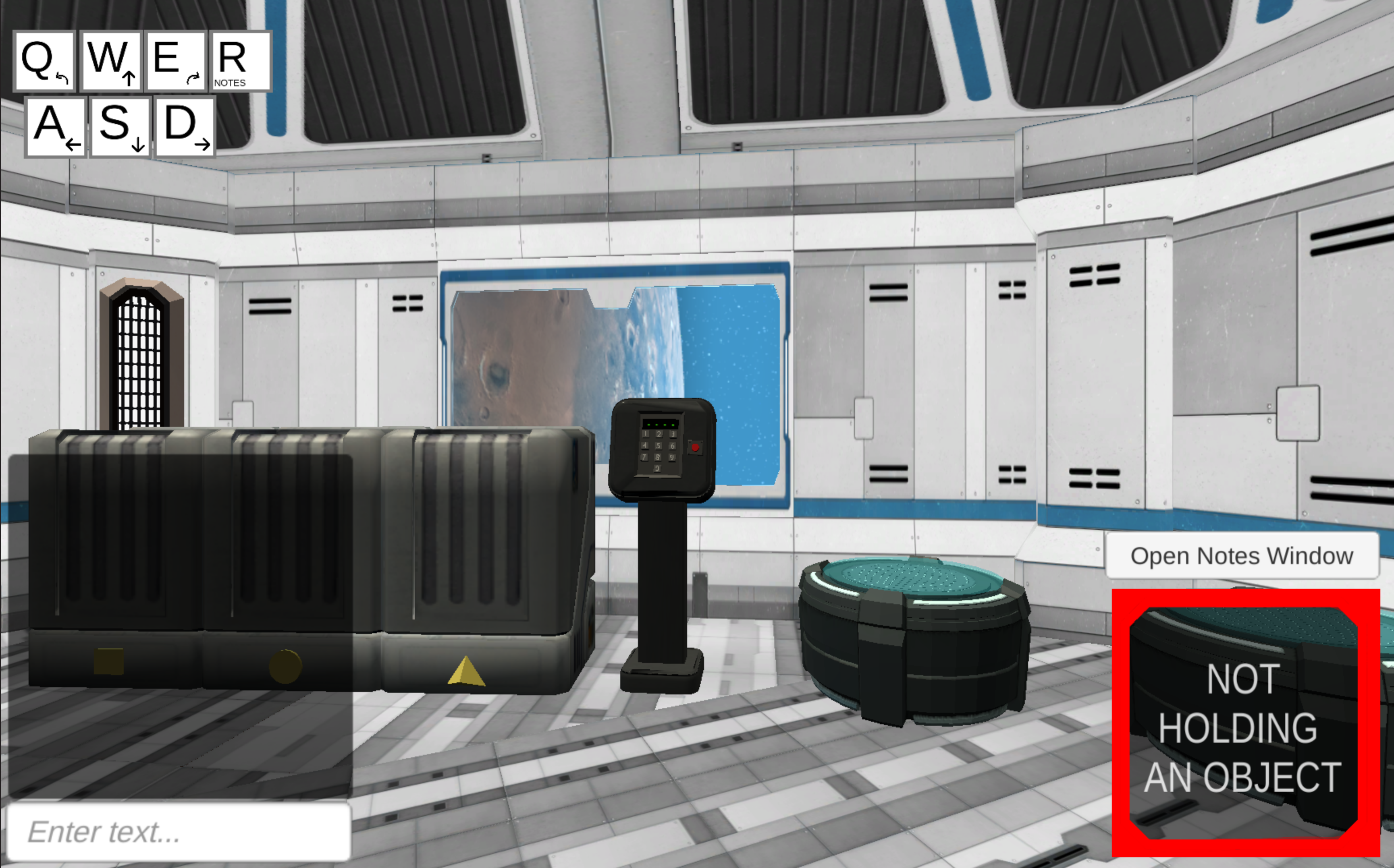}
	\caption{First-person graphical user interface (GUI) used by participants to tele-operate the robot in the study. A message box for communicating with the Commander is in the bottom left, and an ``inventory'' showing the currently-held object is in the bottom right.}
	\label{fig:client_view}
\end{figure}

\begin{figure}[t]
	\centering
	\includegraphics[width=\linewidth]{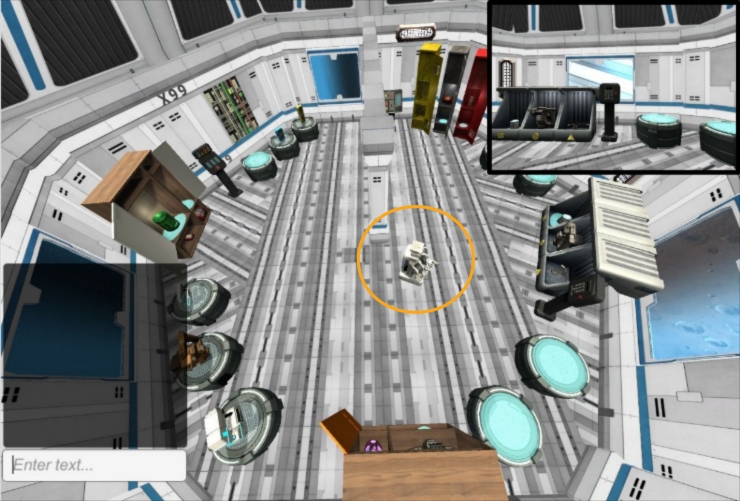}
	\caption{Interface used by the human confederate playing the Commander. The main view shows the robot (circled in orange) in the environment. The top right corner shows the first-person view seen by the human participant playing the robot. The bottom left corner shows a message box for text dialogues with the participant.}
	\label{fig:confederate_view}
\end{figure}

 \subsection{Collaborative Tool Organization Task}
In the task, the human-controlled robot was placed in a virtual spacecraft (see Figure~\ref{fig:client_view}). The task was to organize six tools (among 12 distractors) scattered around the spacecraft -- an activity that is relevant for current and future space robotics applications \cite{bualat2015astrobee}. The tools had to be placed in the proper container (including crates, cabinets, and lockers), some of which were locked and required learning specialized procedures to open. The tools all had fictitious names to ensure that participants were unfamiliar with them, and the tools varied along a number of feature dimensions including color, shape, size, texture, symbol, and pattern. To facilitate the learning process, people could ask questions of a remotely-located Commander (who was played by a human study confederate; see Figure~\ref{fig:confederate_view}) in a live text-based dialogue. 

To explore the effects of several dialogue-level factors, we manipulated \textit{Speaker Initiative} and \textit{Instruction Granularity}, as these have been shown to be relevant for human-agent dialogue \cite{baraglia2016initiative,gervits2016team,marge2020lets}. The Commander took the initiative and gave scripted instructions for half the participants (\textit{Commander Initiative} or \textit{CI}) and only responded to questions in the other half (\textit{Robot Initiative} or \textit{RI}). Half the trials for each participant involved high-level granularity (``The sonic optimizer goes in the secondary cabinet, shelf A") and half involved low-level granularity (``Move to locker Z" $\to$ ``Pick up the sonic optimizer from the top shelf" $\to$ ``Move to the secondary cabinet" $\to$ ``Place the sonic optimizer on shelf A"). In all conditions, the confederate responded to clarification requests with a set policy which generally only provided minimal information. 

\subsection{Interactive Study Platform}
The study was run on Amazon Mechanical Turk (MTurk), which was used for recruitment, questionnaires, and linking to the study environment. To support the proposed study, we developed an infrastructure that enabled interactivity between participants and the experimenter. The environment  was developed in Unity 3D and built in WebGL (a browser-based graphics library)\footnote{This WebGL setup was ideal for MTurk since it rendered directly in the browser and could be linked to from the study page.}. Photon Unity Networking (PUN) was used to support communication between participants and the experimenter and also for synchronizing objects between both views. We used a Willow Garage PR2 robot model and allowed participants to tele-operate the robot directly using the keyboard.

Twenty-two participants recruited from MTurk performed the task. Eleven participants were female, and the average age was 36.8 $\pm$ 7.14. All participants were native English speakers from US zip codes. Participants volunteered by clicking a link on the MTurk page. They then read detailed instructions and performed a tutorial to ensure that they understood the controls and instructions. The tutorial involved a simplified version of the main task with a live experimenter and four simple objects to place (different from the task stimuli). Following successful completion of the tutorial, the task was then performed, which generally took 30-45 minutes. Participants were paid \$10 for completing the study with a possible additional \$2 performance bonus. Video data was recorded of the robot movement and action in the environment, and a transcript of the dialogue was logged. The following measures were taken: \textit{task performance} based on the percentage of the six task-relevant objects placed correctly, \textit{task duration} based on how long it took to complete the task, \textit{questions / total utterances} which indicates the proportion of questions in the dialogue, and \textit{proportion of question types} based on the scheme described in Section~\ref{sec:scheme}.

\section{HuRDL Corpus Overview}
\label{sec:scheme}

\begin{figure}[t]
	\centering
	\includegraphics[width=\linewidth]{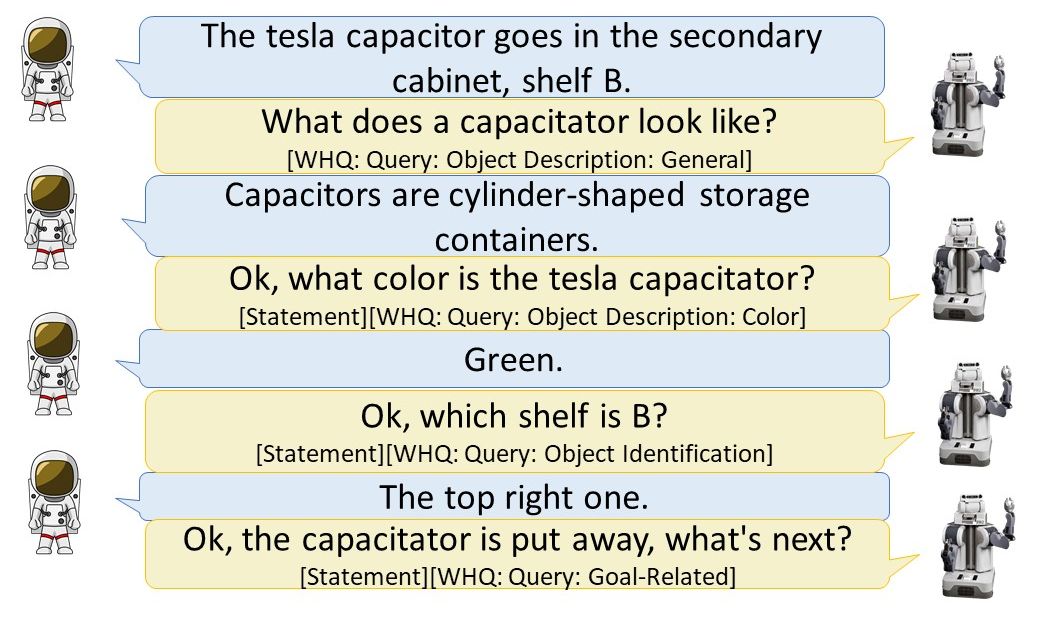}
	\caption{Example dialogue showing a single exchange along with annotations of the participant utterances.}
	\label{fig:dialogue}
\end{figure}

The HuRDL corpus contains twenty-two dialogues with a total duration of 13 hours. It contains a total of 1122 participant utterances, 760 of which are questions. Each dialogue has a mean of 51 participant utterances, 34 of which are questions. The mean score on the task is 77.3\% $\pm$ 24\% and the average duration is 35.2 $\pm$ 7 min. An example dialogue (with annotation) is shown in Figure~\ref{fig:dialogue}.

To analyze question types, two annotators labeled the twenty-two dialogues as described below.  
The annotators began by using one of the dialogues to develop an annotation scheme for the form and clarification type categories described below; this scheme was then refined by performing a consensus annotation on a second dialogue.  
Inter-rater reliability was calculated by having both raters annotate the same four dialogues. Overall, there was 82.2\% raw agreement between annotators, with a Cohen's $\kappa=.79$. For non-statement utterances there was 82.9\% raw agreement with Cohen's $\kappa=.81$. 

\subsection{Annotations: Utterance Forms}

First, utterances were labeled with their \textit{form}, which was one of several categories: \textit{yes/no-questions (YNQs)} are questions that elicit a yes or no response; \textit{alternative questions (AQs)} are questions that present a list of options; \textit{wh- questions (WHQs)} ask who, what, where, when, why, which, or how; and Statements are non-questions.

The annotators found that of this corpus' utterances, 15.3\% are YNQs, 2.8\% are AQs, 49.6\% are WHQs, and 35\% are Statements.  These add up to more than 100\% because some complex utterances contained multiple question forms, or contained both a statement and a question. 

\subsection{Annotations: Clarification Types}

Second, utterances were labeled with their \textit{clarification type} using the types shown in Table~\ref{tab:question_types}.  
By adding question content, this approach expands on previous utterance taxonomies, such as the one presented by \citet{rodriguez2004form}.  

The annotators noted that the clarification type annotations were guided by (but not universally determined by) the form annotations: 
YNQs and AQs tended to be confirmation questions, and WHQs tended to be queries.  
However, the corpus contains several interesting exceptions to this.
For example, consider the corpus utterance: ``shelf D I assume is the bottom right one.''  Although the utterance form is a statement, the utterance is an indirect speech act  \cite{searle-indirectSpeechActs-sas1975} functioning as a question that is seeking to confirm an object according to its location, and doing so with a spatial reference.

As shown in Table~\ref{tab:question_types}, clarification types can either be confirmation questions or queries.  
Confirmation questions can be one of three main classes, either confirming an object based on its location or a feature, or confirming an action.  Each of these has additional sub-classes further specifying the confirmation.  
Queries can be one of several different classes, some of which are non-confirmation questions related to reference resolution, and some of which are requests for task-related instruction (either about opening lockers, or asking the next step in the task).  Several of these also have additional sub-classes. 

\begin{table}[t]
	\caption{Distribution of Clarification Type Annotations Per Total Utterances.}
	\resizebox{\columnwidth}{!}{
		\begin{tabular}{llc}
			\multicolumn{1}{c}{\textbf{Type}}    & \textbf{ \%} & \textbf{Example}                          \\ \hline
			\rowcolor{gainsboro}\textbf{\textsc{Confirmation Questions}}                     &        & \multicolumn{1}{l}{\textit{}}             \\
			\hspace{3mm}\textit{Confirm Object by Location:} & 5.9         & \multicolumn{1}{l}{}                      \\
			\hspace{6mm}Spatial                              & \hspace{3mm}3.8         & \textit{The one on the left?}             \\
			\hspace{6mm}Proximity                            & \hspace{3mm}0.09        & \textit{The nearby one?}                  \\
			\hspace{6mm}Landmark                             & \hspace{3mm}1.6         & \textit{The one next to the wall?}        \\
			\hspace{6mm}Deictic Action                       & \hspace{3mm}0.36        & \textit{The one I'm holding?}             \\
			\hspace{6mm}Other                                & \hspace{3mm}0.09        & \textit{On the second half?}              \\
			\hspace{3mm}\textit{Confirm Object by Feature:}  & 8.9         & \multicolumn{1}{l}{}                      \\
			\hspace{6mm}Size                        & \hspace{3mm}0.62        & \textit{You mean the tall one?}           \\
			\hspace{6mm}Shape                                & \hspace{3mm}0.62        & \textit{The narrow one?}                  \\
			\hspace{6mm}Color                                & \hspace{3mm}3.7         & \textit{The green one?}                   \\
			\hspace{6mm}Pattern                              & \hspace{3mm}0.36        & \textit{The striped one?}                 \\
			\hspace{6mm}Symbol                               & \hspace{3mm}0.45        & \textit{The one with a circle?}           \\
			\hspace{6mm}Hybrid                               & \hspace{3mm}1.78        & \textit{The green cylinder?}              \\
			\hspace{6mm}Comparison                           & \hspace{3mm}0.62        & \textit{The one that looks like a snake?} \\
			\hspace{6mm}Other                                & \hspace{3mm}0.80        & \textit{The one with numbers on it?}      \\
			\hspace{3mm}\textit{Confirm Action:}             & 4.0         & \multicolumn{1}{l}{\textit{}}             \\
			\hspace{6mm}General                              & \hspace{3mm}0.27        & \textit{Did I do it right?}               \\
			\hspace{6mm}Task-Related                & \hspace{3mm}3.6         & \textit{Does the block go in the locker?} \\
			\hspace{6mm}Other                                & \hspace{3mm}0.18        & \textit{Does this light up?}              \\
			\rowcolor{gainsboro}\textbf{\textsc{Queries}}                      &        & \multicolumn{1}{l}{\textit{}}             \\
			\hspace{3mm}\textit{Object Description:}         & 10.9        & \multicolumn{1}{l}{\textit{}}             \\
			\hspace{6mm}General                              & \hspace{3mm}4.8         & \textit{What does it look like?}          \\
			\hspace{6mm}Size                        & \hspace{3mm}0.18        & \textit{What size is it?}                 \\
			\hspace{6mm}Shape                       & \hspace{3mm}0.18        & \textit{What shape is it?}                \\
			\hspace{6mm}Color                                & \hspace{3mm}5.3         & \textit{What color is it}                 \\
			\hspace{6mm}Pattern                              & \hspace{3mm}0.27        & \textit{What pattern is it?}              \\
			\hspace{6mm}Symbol                               & \hspace{3mm}0.18        & \textit{What symbol is on it?}            \\
			\hspace{3mm}\textit{Location-Related}            & 12.3        & \textit{Where is that one?}               \\
			\hspace{3mm}\textit{Object Identification}       & 6.1         & \textit{Which one is that?}               \\
			\hspace{3mm}\textit{Object Naming}               & 0.98        & \textit{What is the small one called?}  \\
			\hspace{3mm}\textit{Goal-Related}                & 9.1         & \textit{What's next?}                     \\
			\hspace{3mm}\textit{Request Teaching:}           & 11.2        & \textit{}                                 \\
			\hspace{6mm}General                     & \hspace{3mm}3.3         & \textit{How do I open lockers?}           \\
			\hspace{6mm}Target                               & \hspace{3mm}7.9         & \textit{What's the code for crate 3?}     \\
	\end{tabular} }
	\label{tab:question_types}
\end{table}

\section{Results and Discussion}
\label{sec:results}
In our analysis of the corpus, we discovered several key findings about how people ask questions under uncertainty.

\subsection{Question Types}
In terms of utterance forms, a one-way MANOVA showed significant differences between the mean proportion of the three main utterance forms (collapsed across all conditions) by total participant utterances, \textit{F}(2,63) = 33.98, \textit{p} $<$ .001. Post-hoc tests using the Bonferroni correction revealed that YNQs were the least frequent, followed by Statements, and then WHQs; \textit{p}s for all comparisons $<$ .05. Given that half of all utterances were WHQs, this finding suggests that WHQs are key questions used by people to reduce uncertainty in this domain.

In terms of clarification types, queries were by far the most common, accounting for 73\% of participant utterances. Interestingly, we found a strong negative correlation between \textit{Location-Related} queries / total questions and task performance in the CI condition, \textit{r}(8) = -.720, \textit{p} $<$ .05. That is, the more location-related questions people asked the worse they performed on average. This correlation could reflect ineffective questions. For example, the experimenter did not know where an object was located, so questions such as ``Where is X?'' were generally ineffective.

Compared to the corpus analysis from \citet{rodriguez2004form} in which 52\% of clarification requests were related to referential ambiguity, in our corpus this was about 75\%. In their corpus, 45\% of response utterances were YN \textit{answers} (suggesting a similar proportion of YNQs), whereas in ours, only 15\% of the questions were YNQs. This was likely a result of the novel objects in our task, which led to more queries. Compared to the corpus analysis in \citet{purver2003means}, our results indicate a large proportion of \textit{non-reprise clarifications}, i.e., explicit questions that do not echo or repeat the instruction. We also found fewer disfluencies in our analysis due to it being written communication. The few observed ones were mostly typos and fragments. Finally, \citet{cakmak2012designing} found that 82\% of all questions in their learning task were feature-based, whereas we observed about 20\% of questions in this category. This can be attributed to differences in task domain and participant familiarity with the environment.  

\subsection{Design and Research Implications}\label{sec:applying-results}

This corpus analysis provides evidence that human dialogue strategies to manage uncertainty can be used to inform the development of real-time, online learning algorithms for agents in situated interaction. Our results directly inform the development of such algorithms in several ways. First, they outline the distribution of question forms and types that people used to manage uncertainty. This distribution (see Table~\ref{tab:question_types}) serves as a guideline about which kinds of questions to use, especially when encountering specific kinds of uncertainty. For example, if multiple objects have the same color, an agent can generate a color query to disambiguate. Second, the results capture the surface form of the questions, i.e., how they were realized. This enables the corpus to serve as a training set for data-driven dialogue systems that can learn to generate questions based on input instructions and their own uncertainty representations. 

While the human data may serve as a good guideline for agent clarification, it is important to acknowledge the limitations of applying the results too directly. For example, people tended to ask about features that were (1) salient and (2) interpretable. Since salience for agents is likely different than for humans, the content of their queries should adjust accordingly. Moreover, though general object descriptions and object identifications were used by humans in the task, they should perhaps be limited by agents since they may not have the perceptual capabilities to interpret the response to general questions such as ``What does it look like?" Instead, feature-based queries that the agent can interpret may be more effective. Prior work in active learning has highlighted the benefit of feature queries (e.g., \citet{bullard2018towards}), however the present work is complementary to such approaches in that it serves as an empirical basis for the form and content of questions that robots should ask once the learning algorithms have determined the missing information. 

Moving forward, the HuRDL corpus has utility as a test bed for further exploration into clarification requests in domains with high uncertainty across modalities. Future work will explore dialogue strategies (i.e., patterns of question types) used by different participants and compare their effectiveness. Additional corpus analysis can investigate other factors that influence question generation including the effects of time pressure, workload, and object properties. Moreover, the video analysis can reveal the visual input that people had access to and the influence that this had on question generation.

It is important that future work apply our annotation scheme to different task domains to establish generalizability. While our task emphasized uncertainty of novel entities to elicit questions, other tasks may not. As a result, we do not expect such a high frequency of questions in other kinds of tasks, nor do we expect the same distribution of question types. The scheme, however, should capture the scope of questions used in a broad range of situated learning tasks since the categories represent general properties by which objects can be identified and distinguished from one another.

\section{Conclusion}
\label{sec:conclusion}

To investigate the problem of how agents can most effectively ask questions in a situated interaction, we analyzed dialogue data from the HuRDL corpus that we collected. The task involved uncertainty across multiple modalities and led to a variety of clarification questions to manage this uncertainty. We categorized these questions in a novel scheme and used it to annotate the corpus. Analysis of question types showed that people used a high frequency of WH-questions, and that these were targeted at learning object features and locations, object task-relevance, goals, and procedural knowledge. These patterns were influenced by dialogue-level factors such as speaker initiative and instruction granularity. Given these results, we presented guidelines to inform automated approaches to effective question generation, which will help make situated agents more resilient in uncertain environments. Future work will develop algorithms for clarification based on the question types and dialogue strategies identified in this corpus.


\section*{Acknowledgments}
This research was sponsored by the Basic Research Office of the U.S. Department of Defense with a Laboratory University Collaboration Initiative Fellowship awarded to MM. The authors would like to thank Genki Kadomatsu and Dean Thurston for their contributions to the study platform.

\bibliographystyle{acl_natbib}
\bibliography{refs}

\end{document}